\documentclass[runningheads]{llncs}
\usepackage{graphicx}
\usepackage[utf8]{inputenc}
\usepackage[english]{babel}
\usepackage{amsmath}
\usepackage{xcolor}
\usepackage{subcaption}
\graphicspath{{./images/}}
\DeclareGraphicsExtensions{.pdf,.jpeg, .png, jpg}

\newcommand{\hide}[1]{}

\begin{document}
	\title{Learning advisor networks for noisy image classification}
	
	%	\author{Anonymous submission}
	
	\author{Simone Ricci \orcidID{0000-0001-9838-6076} \and
		Tiberio Uricchio \orcidID{0000-0003-1025-4541} \and
		Alberto Del Bimbo \orcidID{0000-0002-1052-8322} 
	}
	\authorrunning{S. Ricci et al.}
	\institute{
		University of Florence, Italy \\
		\email{\{name.surname\}@unifi.it}
	}
	\maketitle              % typeset the header of the contribution
	\begin{abstract}
		
		In this paper, we introduced the novel concept of advisor network to address the problem of noisy labels in image classification. Deep neural networks (DNN) are prone to performance reduction and overfitting problems on training data with noisy annotations. Weighting loss methods aim to mitigate the influence of noisy labels during the training, completely removing their contribution. This discarding process prevents DNNs from learning wrong associations between images and their correct labels but reduces the amount of data used, especially when most of the samples have noisy labels. Differently, our method weighs the feature extracted directly from the classifier without altering the loss value of each data. The advisor helps to focus only on some part of the information present in mislabeled examples, allowing the classifier to leverage that data as well. We trained it with a meta-learning strategy so that it can adapt throughout the training of the main model. We tested our method on CIFAR10 and CIFAR100 with synthetic noise, and on Clothing1M which contains real-world noise, reporting state-of-the-art results.
		
	\end{abstract}
	\section{Introduction}
	Modern image classification systems are based on using deep neural network models that are trained on a huge number of labeled images \cite{krizhevsky2012imagenet}. Due to the extreme cost of labeling such an amount of images and difficulty in covering many concepts, researchers recently have looked into methods that generate labels automatically. One significant line of research exploits available labeled images from non-experts (e.g. from social networks, online stores) that can be easily retrieved in large quantities but may have been mislabeled \cite{algan2021image}.
	
	Deep neural networks typically consist of a large number of parameters that are highly shared among feature dimensions and states, enabling flexibility in learning different tasks and classes. This flexibility has the advantage to lead to strong discriminative models unless data annotations are corrupted by noise, leading to performance reduction and overfitting problems \cite{jiang2020beyond}. Recent methods tried to address the problem by using curriculum learning \cite{bengio2009curriculum}, directly estimating the labels noise in the set \cite{hendrycks2018using}, or measuring the confidence of the network during training \cite{kumar2010self}, also using another co-trained network \cite{han2018co}. The idea was usually to understand mislabeled samples out of distribution and reduce their influence on the learning by dampening their loss or decreasing their impact directly from the training set.
	
	In this paper, we proposed a meta-learning approach to address the problem of noisy labels in image classification based on an advisor network, developed to help the classifier. While a standard image classification model is trained, the advisor network observes the main network activations and adjusts features at training time when noisy label images are identified as input. This allows the classifier model to get information even from mislabeled samples where some noise structure is present. We only retained the main model as the final classifier, while the advisor was discarded. Unlike the teacher-student paradigm, the advisor network was not trained to solve the image classification task, but only to help the learning process of the classifier model by its altering activations.   
	
	In summary, our contribution is:
	\begin{itemize}
		\item We propose the use of an advisor network, i.e. the use of an additional network at training time, learned by meta-learning, that can adjust activations and gradient of the main network that is being trained.
		\item We develop such concept for the task of image classification, allowing the training of an image classification network in presence of artificial label noise.
		\item We test our approach in presence of artificial label noise and on a popular noisy dataset, obtaining state-of-the-art performance.  
	\end{itemize}
	
	\section{Related works}
	
	\subsection{Noisy training labels}
	
	Numerous works deal with the problem of noisy labels in training data. It has been shown that the performance of machine learning systems degrades in the presence of label noise \cite{nettleton2010study}, \cite{pechenizkiy2006class}. A first solution involves a loss correction to mitigate the effect of mislabeled samples on the classifier network. For example GLC \cite{hendrycks2018using}, Reed \cite{reed2014training}, M-correction \cite{arazo2019unsupervised}, F-correction \cite{goldberger2016training} and S-adaptation \cite{patrini2017making} estimated the matrix of corruption probabilities used to change the wrong labels to the correct ones. Instead, \cite{tanaka2018joint}, \cite{ma2018dimensionality}, \cite{yi2019probabilistic} modeled the annotations noise distribution linearly combining the output of the network and the noisy label to estimate true labels. Another different approach was assigning a weight to each sample. A lower weight value avoids the contribution of that sample to the training of the network. In this way, it is possible to assign low values to mislabeled examples and high values to correct ones. MentorNet \cite{jiang2018mentornet} and MentorMix \cite{jiang2020beyond} found the latent weights with data-driven curriculum learning and the student-teacher paradigm. Other contributions include data augmentation strategies like Mixup \cite{zhang2018mixup}, Advaug \cite{cheng2020advaug} and DevideMix \cite{li2019dividemix}. Differently from these methods, we modified the network activation using an advisor instead of the loss value.
	
	\subsection{Meta learning}
	
	There are methods \cite{azadi2015auxiliary}, \cite{vahdat2017toward}, \cite{veit2017learning}, \cite{li2017learning} that needs supplemental clean label to handle the noise. This assumption of clean data is also true for a solution that exploits the Meta-learning paradigm. It consists of the use of machine learning algorithms to assist the training and optimization of other machine learning models. Meta-learning \cite{ren2018learning}, \cite{shu2019meta}, \cite{li2019learning}, \cite{wang2020training} had used to address the noisy labels problem. With small clean validation data, the meta-model learns how to correct the biased training labels. For example, L2R \cite{ren2018learning} weighed each example giving less importance to the mislabeled samples. MLNT \cite{li2019learning} simulated regular training with synthetic noisy labels. MW-Net \cite{shu2019meta} learned an explicit weighting function that can be easily adapted to different types of annotations noise. MLC \cite{wang2020training} estimated the label noise transition matrix. Contrary to all aforementioned meta-learning solutions, our method does not act by directly modifying the loss of the neural network. We applied a meta-attention layer inside a neural network. The weights of the attention are learned by the advisor network. In this way, the mislabeled data can be leveraged to improve the overall performance of the main model.
	
	\section{Method}
	
	\subsection{Task}
	
	In this paper, we developed a method that can handle images with noisy labels when training networks for image classification. We started from the idea that also a mislabeled example contains information that can contribute to a greater generalization of the network. The model should concentrate only on some convenient parts of these data. Our idea was to exploit the attention mechanism to enhance the useful parts of the information and lower the rest. We made use of an auxiliary advisor network that learns automatically a function that weighs the features extracted from a DNN during its training. This advisor network should be aware of the state of the main model and the meta-learning training solves this constraint. Our method Meta Feature Re-Weighting (MFRW) acts like a meta-attention layer. Different from weighting loss methods that tend to completely remove the influence of mislabeled examples during the training our MFRW can take advantage of them.
	
	We first introduce meta-learning basics and formulation typical of methods that learn robust deep neural networks from noisy labels. Then in Section \ref{mfrw}, we explain our method showing the architecture of the whole process.
	
	\subsection{Meta learning for Noisy Image Classification}\label{back}
	
	In general meta-learning (ML) is referred to the process of improving a learning algorithm over multiple learning episodes, also called commonly learning to learn. Usually, ML is divided into two learning algorithms: an inner (or base) algorithm that solves a task, such as images classification, defined by a training dataset and objective function; an outer (or upper/meta) algorithm that updates the inner one, such that the main model it learns improves an outer objective function. ML was applied to solve the problem of noisy labels in training data \cite{ren2018learning}, \cite{shu2019meta}. We introduce the symbols useful for understanding ML in this particular setting and the three basic steps into which the entire learning process is divided.
	
	Let $D^{train} = \{x_i^{tra}, y_i^{tra}\}^N_{i=1}$ be the noisy annotated training set, where N is the total number of examples, composed of an image $x_i$ and the correspondent one-hot label $y_i$ over $c$ classes. In general if we have a deep neural network (DNN) model $\Phi(\cdot; w)$, where $w$ are its parameters and $ \hat{y} = \Phi(x; w)$ is its prediction on the input image $x$, we can obtain the optimal parameters $w^*$ by minimizing the softmax cross-entropy loss $\ell(\hat{y}, y)$ on the training set $D^{train}$. ML, applied to the Noisy Image Classification task, requires the presence of an additional verified dataset. This validation set $D^{val} = \{x_j^{val}, y_j^{val}\}^M_{j=1}$ is much smaller than the training set, $M \ll N$.
	
	In \cite{shu2019meta} a meta-model was used to implement the ML process. A multilayer perceptron network with only one hidden layer learns how to weigh each training example. Let $\Psi(\cdot;\theta)$, parameterized by $\theta$, be the meta-model that maps a loss value to a scalar weight. In this case, the optimal parameters $w^*$ are derived using the following weighted loss:
	
	\begin{equation}\label{MW1}
		w^* (\theta) = \underset{w}{\mathrm{argmin}} \frac{1}{N} \sum_{i=1}^{N} \mathcal{V}_i^{tra}(\theta)\mathcal{L}_i^{tra}(w)
	\end{equation}
	with $\mathcal{V}_i^{tra}(\theta) = \Psi(\mathcal{L}_i^{tra}(w);\theta)$ as the weight predicted by the meta model for the $i$-th training example. Instead the meta model is trained minimizing the validation loss:
	
	\begin{equation}\label{MW2}
		\theta^* = \underset{\theta}{\mathrm{argmin}} \frac{1}{M} \sum_{j=1}^{M} \mathcal{L}_j^{val}(w^*(\theta))
	\end{equation}
	where $ \mathcal{L}_j^{val}(w^*(\theta)) = \ell( \Phi(x_j^{val};w^*(\theta),y_j^{val}  ) )$ is the loss for the $j$-th validation example.
	
	Equations Eq. (\ref{MW1}) and Eq. (\ref{MW2}) can be minimized alternating optimization via gradient descent. One solution that ensures the efficiency of the algorithm and its convergence \cite{shu2019meta} adopts an online strategy to update $\theta$ and $w$ through a single optimization loop, which is divided into three main steps.
	
	The first step is called Virtual-Train because the original DNN will not be updated and the optimization is carried out on a virtual model that is the copy of the original one. Consider the $t$-th iteration and associated mini batches $\mathcal{B}^{train} = \{ (x_i^{tra},y_i^{tra}) \}_{i=1}^n$ and $\mathcal{B}^{val} = \{ (x_j^{val},y_j^{val}) \}_{j=1}^m$, where $n$ and $m$ are the size of mini-batch respectively. The virtual update can be derived by:
	\begin{equation}\label{MW3}
		\hat{w}(\theta) = w - \alpha \frac{1}{n}\sum_{i=1}^{n}\mathcal{V}_i^{tra}(\theta)\nabla_w\mathcal{L}_i^{tra}(w)
	\end{equation}
	where $\alpha$ is the learning rate for the DNN and $w$ is its parameter at the current iteration. Then there is the Meta-Train step, where given the optimized virtual model the meta model is updated by:
	\begin{equation}\label{MW4}
		\theta' = \theta - \beta \frac{1}{m}\sum_{j=1}^{m}\nabla_\theta\mathcal{L}_i^{val}(\hat{w}(\theta))
	\end{equation}
	with $\beta$ the learning rate for the meta model. In last step, Actual-Train, the base DNN model is optimized taking into account the previously updated meta model.
	\begin{equation}\label{MW5}
		w' =  w - \alpha \frac{1}{n}\sum_{i=1}^{n}\mathcal{V}_i^{tra}(\theta')\nabla_w\mathcal{L}_i^{tra}(w)
	\end{equation}
	$w'$ becomes the $w$ in Eq. (\ref{MW3}) for the $(t+1)$-th iteration.

	\subsection{Meta Feature Re-Weighting (MFRW)}\label{mfrw}
	
	Attention for a DNN is a mechanism that tries to mimic the cognitive attention of the human brain. It intensifies the important parts of the input and reduces the rest. In Meta Feature Re-Weighting (MFRW) the attention is applied with a Hadamard product between the feature extracted from a DNN and a vector of weights automatically learned from a meta-model. In order to get this, we separated the main model $\Phi(\cdot; w)$ in two-part: the backbone $\Phi_b(\cdot; w_b)$, that has an image $x$ as input and gives out a feature vector $f$, and the classifier $\Phi_c(\cdot; w_c)$, that has $f$ as input and a probability score vector $s$ as output. In this way, it was possible to manipulate the feature $f$ directly with our meta-model $\Psi$.
	
	The meta-model takes two different inputs $\Psi(f,\mathcal{L})$ and gives back a vector of weights $W_f$. The first input $f$ is the feature extracted from the backbone $\Phi_b$ relative to the example $x$. This is important for the meta-model because it makes the $W_f$ strictly connected to the feature that needs to be modified. The other input is the loss $\mathcal{L}$ of the example $x$ calculated from the prediction obtained by the main full model $\Phi$. This gives the meta-model information about how much $x$ is a ``hard" or an ``easy" example for the main model. The two inputs together let the meta-model differentiate a feature belonging to a noisy $x$ from the one related to a correct $x$. In dot-product attention the multiplication is done element-wise, so the $W_f$ has to be of the same size as $f$, and its values must be in the range $\in (0,1)$.
	
\begin{figure}
		\centering
		\includegraphics[width=0.85\linewidth]{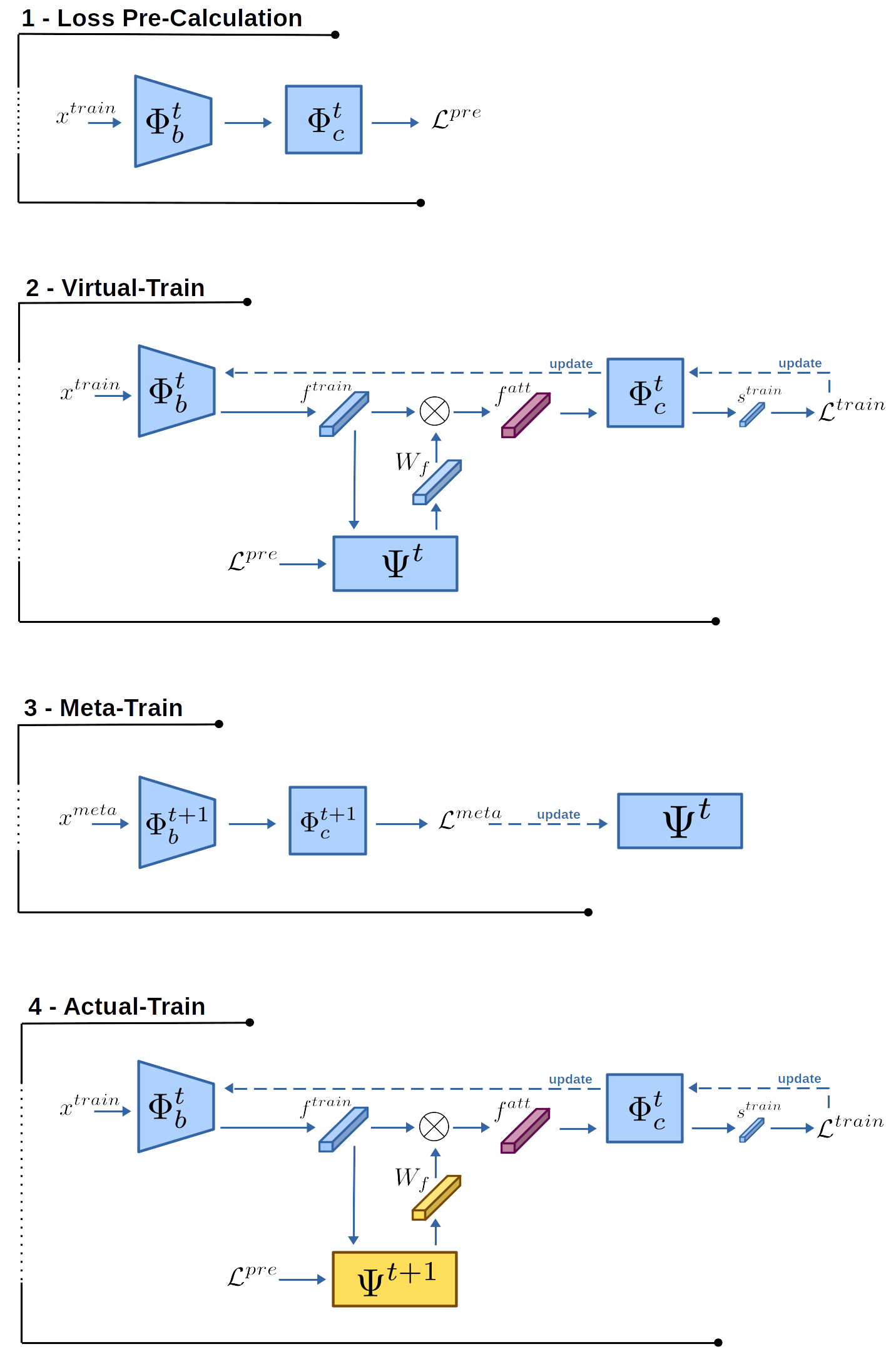}
		\caption{Illustration of an iteration of the proposed Meta Feature Re-Weighting (MFRW) method. Each iteration is divided into four steps. First, a Loss Pre-Calculation is performed to calculate in advance the loss $\mathcal{L}^{pre}$ value of the training batch $x^{train}$. The second step is the Virtual-Train, where a clone of the main model is virtually updated. Here the meta-model modifies the feature of the main model by multiplying it with a vector of weights. The purple color indicates the weighted features. The third step shows the Meta-Train process. With a meta batch of clean example $x^{meta}$ the meta-model is updated minimizing the loss $\mathcal{L}^{meta}$ given by the previous virtually updated network. In the last phase Actual-Train, the real main model is trained with the meta-model optimized (yellow color)}
		\label{fig:schema1}
	\end{figure}

	MFRW is divided into 4 main phases for each iteration. Figure \ref{fig:schema1} shows the overall process of our method divided by step. We put our method at the $t$-th iteration and we will describe each step to reach the $(t+1)$-th.
	
	Our method needs an additional initial phase Loss Pre-Calculation respect to \cite{shu2019meta} and what is described in \ref{back}. We must calculate in advance the value of loss $\mathcal{L}^{pre}$ related to the training batch $x^{train}$. This is done at the beginning to obtain a loss value dependent on the original feature and not on the weighted one. It is not an expensive step because it is a direct loss inference, without gradient calculation.
	
	The second step is the Virtual-Train. Here $\Phi_b^t$ and $\Phi_c^t$ are temporary clone of the original ones. The train batch $x^{train}$ is passed in $\Phi_b^t$ to obtain the features $f^{train}$. Then $f^{train}$ goes inside $\Psi^t$ with the relative loss values $\mathcal{L}^{pre}$ to get the vector of weights $W_f$. We multiplied element-wise $f^{train}$ with $W_f$ to get a new feature vector with attention $f^{att}$. This is given to $\Phi_c^t$ to obtain the score $s^{train}$ and then the correspondent loss $\mathcal{L}^{train}$. We now virtually update $\Phi_b^t$ and $\Phi_c^t$ parameters, but not the one of $\Psi^t$.
	
	Like \cite{shu2019meta} we have a clean and balanced meta dataset that will be used to train the meta-model $\Psi$ in the third step Meta-Train. Here we have $\Phi_b^{t+1}$ and $\Phi_c^{t+1}$ virtually updated from the step before. Now we pass a meta batch $x^{meta}$ through them in order to get a validation loss $\mathcal{L}^{meta}$. Then $\Psi^t$ is updated minimizing $\mathcal{L}^{meta}$. In this way, the meta-model is optimized to help the main model minimize its error on clean data. Here the optimization takes into consideration also the previous Virtual-Train, thus this is the most expensive part of the method.
	
	The last phase is the Actual-Train where the original $\Phi_b^t$ and $\Phi_c^t$ are optimized taking into account the updated meta-model $\Psi^{t+1}$.
	
	The meta-model is used only during the training time of the main network. It will be discarded at test time when only the main network is retained as the final model.
	
	\subsection{Meta model architecture}
	
	Our meta-model $\Psi$ has a really simple architecture. The inputs of the network are a feature $f$ and a loss value $\mathcal{L}_x$. Each input is projected in a fixed size embedding space through a separate fully connected layer. Then the embeddings are concatenated and passed to another fully connected layer that projects them into a larger common space. Its size is the sum of the dimension of each previous embedding. Finally, a linear layer is used to pass the data from the common space to a vector with a size equal to the one of the feature $f$, that is given as input. Because the output must be an attention weight in the range $\in (0,1)$ we put a sigmoid activation after the last layer.

	\section{Experiments}
	%alto livello 
	
	%Creating very large datasets comes at a very high cost. To solve this problem, many new datasets are generated via automated systems or from labeling provided by inexperienced annotators (web users) and surrounding texts. This often results in the introduction of noise into the data.
	To demonstrate the effectiveness of our method, we conducted experiments on synthetically generated datasets with controlled noise structure and level. Then we tested its ability to generalize with experiments on a real-world dataset.
	
	%dataset artificial

	\subsection{Datasets}
	
	Following previous work \cite{shu2019meta}, \cite{ren2018learning}, \cite{jiang2018mentornet}, we used CIFAR-10 and CIFAR-100 which are the typical choice to generate synthetic datasets containing different types of noise structures. They are composed of 50K training images and 10K test images of size 32×32. Of the training set, 1000 images with clean labels are randomly selected to create the validation set for meta-training.
	
	In addition to synthetic datasets, there is a collection of data containing real-world noise. Clothing1M \cite{xiao2015learning} is a dataset that is composed of 1 million images of clothing taken from online shopping websites. There are $14$ categories like T-shirt, Shirt, Knitwear, etc. The labels are obtained from the text of the images provided by the sellers and not from expert annotators, that's why there are errors. The validation set of $14k$ clean data is used as the meta dataset. This dataset allowed our strategy to be evaluated as a concrete application for fine-grained classification with noisy training annotations.
	
	\subsection{Implementation details}
	
	We used the same settings for the experiments on CIFAR-10 and CIFAR-100. The backbone was a Resnet-32 trained through SGD with a momentum of $0.9$, weight decay of 5e-4, batch size of $128$, and a starting learning rate of 0.1. The learning rate decreased to its $\frac{1}{10}$ at the 50 epoch and 70 epoch, stopped at the 100 epochs.
	
	With Clothing1M we used as backbone a ResNet-50 pre-trained on ImageNet. It was trained through SGD with a momentum of $0.9$, weight decay of 1e-3, and a starting learning rate of 0.01. The batch size was $32$ and it was preprocessed resizing the image to 256 × 256, cropping the center 224 × 224, and performing normalization. The learning rate was divided by $\frac{1}{10}$ after 5 epochs and stopped at 10 epochs.
	
	In every experiment, the meta-model was optimized with Adam and a learning rate of 1e-4. The embedding space size was set always to $100$.
	
	\subsection{Results}\label{noise}
	
		Flip (or asymmetric) is a noise that is designed to mimic the structure where labels are only replaced by similar classes, e.g. dog$\leftrightarrow$cat. We choose to test our method on that type of noise because it usually appends that the label error could depend on the ambiguity between classes and similar visual patterns \cite{xiao2015learning}. We created a synthetic version of CIFAR-10 and CIFAR-100. The noise ratio was controlled by a parameter $p$, which represents the probability that a clean example is contaminated by noise. In this way we could test our method on different level of noise, from $p = 0.0$ (no noise), to $p=0.8$ (heavy noise).
	
	\begin{table}[!ht]
		\centering
		\caption{\label{flip} Top-1 accuracy on CIFAR10 and CIFAR100 dataset with Flip noise. The backbone used was a ResNet-32. $p$ denotes the different levels of noise. The results for the cited method are reported directly from their original papers. $^\dagger$ indicates the results obtained by our implementation. The first and the second best results are respectively marked in bold and underline.}
		\resizebox{\columnwidth}{!}{
			\begin{tabular}{c|c|c|c|c|c||c|c|c|c|c}
				\hline
				Dataset     & \multicolumn{5}{c||}{Flip CIFAR-10}                                        & \multicolumn{5}{c}{Flip CIFAR-100}                                       \\ \hline
				Noise $p$       & 0.0   & 0.2            & 0.4            & 0.6            & 0.8            & 0.0   & 0.2            & 0.4            & 0.6            & 0.8            \\ \hline
				CrossEntropy \cite{shu2019meta}    & 92.89 & 76.83          & 70.77          & -              & -              & 70.50 & 50.86          & 43.01          & -              & -              \\
				Reed-Hard \cite{reed2014training}  & 92.31 & 88.28          & 81.06          & -              & -              & 69.02 & 60.27          & 50.40          & -              & -              \\ 
				S-Model \cite{goldberger2016training}    & 83.61 & 79.25          & 75.73          & -              & -              & 51.46 & 45.45          & 43.8           & -              & -              \\ 
				Self-paced \cite{kumar2010self} & 88.52 & 87.03          & 81.63          & -              & -              & 67.55 & 63.63          & 53.51          & -              & -              \\ 
				Focal Loss \cite{lin2017focal} & \underline{93.03} & 86.45          & 80.45          & -              & -              & 70.02 & 61.87          & 54.13          & -              & -              \\ 
				Co-teaching \cite{han2018co} & 89.87 & 82.83          & 75.41          & -              & -              & 63.31 & 54.13          & 44.85          & -              & -              \\ 
				D2L \cite{ma2018dimensionality}        & 92.02 & 87.66          & 83.89          & -              & -              & 68.11 & 63.48          & 51.83          & -              & -              \\ 
				Fine-tuning \cite{shu2019meta} & \textbf{93.23} & 82.47          & 74.07          & -              & -              & \textbf{70.72} & 56.98          & 46.37          & -              & -              \\ 
				MentorNet \cite{jiang2018mentornet}   & 92.13 & 86.3           & 81.76          & -              & -              & 70.24 & 61.97          & 52.66          & -              & -              \\ 
				L2RW \cite{ren2018learning}        & 89.25 & 87.86          & 85.66          & -              & -              & 64.11 & 57.47          & 50.98          & -              & -              \\ 
				GLC \cite{hendrycks2018using}        & 91.02 & 89.68          & \underline{88.92}          & -              & -              & 65.42 & 63.07          & \textbf{62.22} & -              & -              \\ 
				MW-net \cite{shu2019meta}     & 92.04 & 90.33          & 87.54          & -              & -              & 70.11 & \underline{64.22}          & 58.64          & -              & -              \\ 
				CrossEntropy$^\dagger$     &   92.33    & 90.56          & 86.25          & 26.67          & 13.58          & 70.18       & \textbf{65.02} & 50.25          & 18.67          & 4.32           \\ 
				MW-net$^\dagger$ \cite{shu2019meta}     &  92.19     & \underline{90.74}          & 87.63          & \underline{42.41}          & \underline{27.19}          & \underline{70.57}      & 64.13          & 51.23          & \underline{19.89}          & \underline{7.42}           \\  \hline
				\textbf{Ours}        &  91.87     & \textbf{91.09} & \textbf{90.26} & \textbf{89.34} & \textbf{82.47} & 68.93      & 63.54          & \underline{59.07}          & \textbf{56.13} & \textbf{20.29} \\ \hline
			\end{tabular}
		}
	\end{table}
	
	Table \ref{flip} shows the accuracy results on the test set of CIFAR-10 and CIFAR-100 datasets with flip label noises. The compared methods are directly cited with the result in their paper. For MW-Net \cite{shu2019meta} and the direct training (CrossEntropy) we report also our reproduced results. %denoting them with $\dagger$. 
	The accuracy gained over the other methods was significant. We can see that at a higher noise rate our result outperforms MW-Net and CrossEntropy by a large margin, indicating the effectiveness of our method on the synthetic Flip noise.  From the results of Table \ref{flip} is possible to notice a limitation of our strategy that occurs when there is no noise ($p = 0.0$) in the training annotations. We obtained worse accuracy values than the training with the classic softmax cross-entropy loss on both CIFAR-10 and CIFAR-100. The advisor network introduces a bias from the distribution of the meta set to the training data. Because the training annotations are completely correct the introduction of this meta bias makes the accuracy a little worse than without.
	
	We introduced also two new noise settings, namely Flip2 and Flip3. The difference from Flip is that the noise is equally distributed over multiple similar classes, two and three respectively. Table \ref{flip2}, \ref{flip3} show respectively the result for noise of type Flip2 and Flip3. We can see how our method performs better than the others, especially in very noisy situations.
	
	\begin{table}[!ht]
		\caption{\label{flip2} Accuracy result on CIFAR10 and CIFAR100 dataset with Flip2 noise. $p$ denotes the different level of noise. $^\dagger$ indicates the results obtained by our implementation. The first and the second best results are respectively marked with bold and underline.}
		\centering
		\begin{tabular}{c|c|c|c|c||c|c|c|c}
			\hline
			Dataset  & \multicolumn{4}{c||}{Flip2 CIFAR-10} & \multicolumn{4}{c}{Flip2 CIFAR-100} \\ \hline
			Noise $p$    & 0.2        & 0.4       & 0.6       & 0.8       & 0.2        & 0.4        & 0.6       & 0.8        \\ \hline
			CrossEntropy$^\dagger$  & \underline{90.71}      & 87.83      & 75.83 & 11.86     & \underline{64.91}      & 57.7       & 36.55     & 7 \\
			MW-net$^\dagger$ \cite{shu2019meta}   & \textbf{90.93}      & \underline{88.83}      & \underline{86.85} & \underline{27.49}     & \textbf{65.37}      & \textbf{59}         & \underline{36.97}    & \underline{7.99}  \\ \hline
			\textbf{Ours}     & 90.66      & \textbf{89.72}      & \textbf{87.75}    & \textbf{73.83}  & 63.07      & \underline{57.96}      & \textbf{45.35}     & \textbf{22.41} \\ \hline
		\end{tabular}
	\end{table}
	
	\begin{table}[!t]
		\caption{\label{flip3} Result for Flip3 noise on CIFAR10 and CIFAR100 dataset. $p$ denotes the different level of noise. $^\dagger$ indicates the results obtained by our implementation. The first and the second best results are respectively marked with bold and underline.}
		\centering
		\begin{tabular}{c|c|c|c|c||c|c|c|c}
			\hline
			Dataset  & \multicolumn{4}{c||}{Flip3 CIFAR-10} & \multicolumn{4}{c}{Flip3 CIFAR-100} \\ \hline
			Noise $p$    & 0.2        & 0.4       & 0.6       & 0.8       & 0.2        & 0.4        & 0.6       & 0.8       \\ \hline
			CrossEntropy$^\dagger$  & 90.13      & 88.44     & 82.31    & 20.34 & \underline{65.29}      & \underline{59.35}      & 44      & \underline{11.07}  \\
			MW-net$^\dagger$ \cite{shu2019meta}   & \textbf{90.56}      & \underline{88.49}     & \underline{85.65}  &\underline{22.69}   & \textbf{65.33}      & \textbf{62.74}      & \underline{45.77}  &10.33   \\ \hline
			\textbf{Ours}     & \underline{90.31}      & \textbf{88.96}     & \textbf{87.73}    & \textbf{75.53} & 62.98      & 59.08      & \textbf{52.28}   &\textbf{25.72}  \\ \hline
		\end{tabular}
	\end{table}

	Table \ref{clothing} shows the results on Clothing1M. As we can see our method outperforms the current state-of-the-art result.

	\begin{table}[!t]
		\centering
		\caption{\label{clothing} Comparison with state-of-the-art methods in test accuracy $(\%)$ on Clothing1M dataset. Results for baselines are copied from original papers.}
		\begin{tabular}{l|c}
			\hline
			Method       & Accuracy (\%) \\ \hline
			CrossEntropy \cite{shu2019meta}          & 68.94         \\
			F-correction \cite{patrini2017making} & 69.84         \\
			JoCoR \cite{wei2020combating}       & 70.30         \\
			S-Model \cite{goldberger2016training} & 70.36         \\
			M-correction \cite{arazo2019unsupervised} & 71.00         \\
			MLC \cite{wang2020training}         & 71.06         \\
			Joint-Optim \cite{tanaka2018joint}  & 72.16         \\
			MLNT \cite{li2019learning}        & 73.47         \\
			P-correction \cite{yi2019probabilistic} & 73.49         \\
			MW-Net \cite{shu2019meta}      & 73.72         \\
			MentorMix \cite{jiang2020beyond}   & 74.30         \\
			FaMUS \cite{xu2021faster} & 74.43 \\
			DivideMix \cite{li2019dividemix} & 74.76 \\
			AugDesc \cite{nishi2021augmentation} & 75.11 \\ \hline
			\textbf{Ours}         & \textbf{75.35}         \\ \hline
		\end{tabular}
	\end{table}
	
	\section{Conclusions}
	
	In this paper, we introduced Meta Feature Re-Weighting (MFRW), which makes use of a novel concept of advisor network to mitigate the problem of training DNNs on corrupted labels. We empirically show the effectiveness of our method on a synthetic and real-world noisy dataset for the classification task. The experimental results demonstrate that the advisor strategy can leverage information present in noisy data helping the main network to achieve a better generalization performance. Our method yields state-of-the-art performance on the Clothing1M dataset. Future research in this area may include adapting the advisor network to different problems than noise, like class imbalance.
	
	\bibliographystyle{splncs04}
	\bibliography{acmart}

\begin{thebibliography}{10}
\providecommand{\url}[1]{\texttt{#1}}
\providecommand{\urlprefix}{URL }
\providecommand{\doi}[1]{https://doi.org/#1}

\bibitem{algan2021image}
Algan, G., Ulusoy, I.: Image classification with deep learning in the presence
  of noisy labels: A survey. Knowledge-Based Systems  \textbf{215},  106771
  (2021)

\bibitem{arazo2019unsupervised}
Arazo, E., Ortego, D., Albert, P., O’Connor, N., McGuinness, K.: Unsupervised
  label noise modeling and loss correction. In: International Conference on
  Machine Learning. pp. 312--321. PMLR (2019)

\bibitem{azadi2015auxiliary}
Azadi, S., Feng, J., Jegelka, S., Darrell, T.: Auxiliary image regularization
  for deep cnns with noisy labels. arXiv preprint arXiv:1511.07069  (2015)

\bibitem{bengio2009curriculum}
Bengio, Y., Louradour, J., Collobert, R., Weston, J.: Curriculum learning. In:
  Proceedings of the 26th annual international conference on machine learning.
  pp. 41--48 (2009)

\bibitem{cheng2020advaug}
Cheng, Y., Jiang, L., Macherey, W., Eisenstein, J.: Advaug: Robust adversarial
  augmentation for neural machine translation. In: Proceedings of the 58th
  Annual Meeting of the Association for Computational Linguistics. pp.
  5961--5970 (2020)

\bibitem{goldberger2016training}
Goldberger, J., Ben-Reuven, E.: Training deep neural-networks using a noise
  adaptation layer  (2016)

\bibitem{han2018co}
Han, B., Yao, Q., Yu, X., Niu, G., Xu, M., Hu, W., Tsang, I., Sugiyama, M.:
  Co-teaching: Robust training of deep neural networks with extremely noisy
  labels. Advances in Neural Information Processing Systems  (2018)

\bibitem{hendrycks2018using}
Hendrycks, D., Mazeika, M., Wilson, D., Gimpel, K.: Using trusted data to train
  deep networks on labels corrupted by severe noise. Advances in Neural
  Information Processing Systems  \textbf{31},  10456--10465 (2018)

\bibitem{jiang2020beyond}
Jiang, L., Huang, D., Liu, M., Yang, W.: Beyond synthetic noise: Deep learning
  on controlled noisy labels. In: International Conference on Machine Learning.
  pp. 4804--4815. PMLR (2020)

\bibitem{jiang2018mentornet}
Jiang, L., Zhou, Z., Leung, T., Li, L.J., Fei-Fei, L.: Mentornet: Learning
  data-driven curriculum for very deep neural networks on corrupted labels. In:
  International Conference on Machine Learning. pp. 2304--2313. PMLR (2018)

\bibitem{krizhevsky2012imagenet}
Krizhevsky, A., Sutskever, I., Hinton, G.E.: Imagenet classification with deep
  convolutional neural networks. Advances in neural information processing
  systems  \textbf{25} (2012)

\bibitem{kumar2010self}
Kumar, M., Packer, B., Koller, D.: Self-paced learning for latent variable
  models. Advances in neural information processing systems  \textbf{23},
  1189--1197 (2010)

\bibitem{li2019dividemix}
Li, J., Socher, R., Hoi, S.C.: Dividemix: Learning with noisy labels as
  semi-supervised learning. In: International Conference on Learning
  Representations (2019)

\bibitem{li2019learning}
Li, J., Wong, Y., Zhao, Q., Kankanhalli, M.S.: Learning to learn from noisy
  labeled data. In: Proceedings of the IEEE/CVF Conference on Computer Vision
  and Pattern Recognition. pp. 5051--5059 (2019)

\bibitem{li2017learning}
Li, Y., Yang, J., Song, Y., Cao, L., Luo, J., Li, L.J.: Learning from noisy
  labels with distillation. In: Proceedings of the IEEE International
  Conference on Computer Vision. pp. 1910--1918 (2017)

\bibitem{lin2017focal}
Lin, T.Y., Goyal, P., Girshick, R., He, K., Doll{\'a}r, P.: Focal loss for
  dense object detection. In: Proceedings of the IEEE international conference
  on computer vision. pp. 2980--2988 (2017)

\bibitem{ma2018dimensionality}
Ma, X., Wang, Y., Houle, M.E., Zhou, S., Erfani, S., Xia, S., Wijewickrema, S.,
  Bailey, J.: Dimensionality-driven learning with noisy labels. In:
  International Conference on Machine Learning. pp. 3355--3364. PMLR (2018)

\bibitem{nettleton2010study}
Nettleton, D.F., Orriols-Puig, A., Fornells, A.: A study of the effect of
  different types of noise on the precision of supervised learning techniques.
  Artificial intelligence review  \textbf{33}(4),  275--306 (2010)

\bibitem{nishi2021augmentation}
Nishi, K., Ding, Y., Rich, A., Hollerer, T.: Augmentation strategies for
  learning with noisy labels. In: Proceedings of the IEEE/CVF Conference on
  Computer Vision and Pattern Recognition. pp. 8022--8031 (2021)

\bibitem{patrini2017making}
Patrini, G., Rozza, A., Krishna~Menon, A., Nock, R., Qu, L.: Making deep neural
  networks robust to label noise: A loss correction approach. In: Proceedings
  of the IEEE conference on computer vision and pattern recognition. pp.
  1944--1952 (2017)

\bibitem{pechenizkiy2006class}
Pechenizkiy, M., Tsymbal, A., Puuronen, S., Pechenizkiy, O.: Class noise and
  supervised learning in medical domains: The effect of feature extraction. In:
  19th IEEE symposium on computer-based medical systems (CBMS'06). pp.
  708--713. IEEE (2006)

\bibitem{reed2014training}
Reed, S., Lee, H., Anguelov, D., Szegedy, C., Erhan, D., Rabinovich, A.:
  Training deep neural networks on noisy labels with bootstrapping. arXiv
  preprint arXiv:1412.6596  (2014)

\bibitem{ren2018learning}
Ren, M., Zeng, W., Yang, B., Urtasun, R.: Learning to reweight examples for
  robust deep learning. In: International Conference on Machine Learning. pp.
  4334--4343. PMLR (2018)

\bibitem{shu2019meta}
Shu, J., Xie, Q., Yi, L., Zhao, Q., Zhou, S., Xu, Z., Meng, D.:
  Meta-weight-net: learning an explicit mapping for sample weighting. In:
  Proceedings of the 33rd International Conference on Neural Information
  Processing Systems. pp. 1919--1930 (2019)

\bibitem{tanaka2018joint}
Tanaka, D., Ikami, D., Yamasaki, T., Aizawa, K.: Joint optimization framework
  for learning with noisy labels. In: Proceedings of the IEEE Conference on
  Computer Vision and Pattern Recognition. pp. 5552--5560 (2018)

\bibitem{vahdat2017toward}
Vahdat, A.: Toward robustness against label noise in training deep
  discriminative neural networks. Advances in Neural Information Processing
  Systems  \textbf{30},  5596--5605 (2017)

\bibitem{veit2017learning}
Veit, A., Alldrin, N., Chechik, G., Krasin, I., Gupta, A., Belongie, S.:
  Learning from noisy large-scale datasets with minimal supervision. In:
  Proceedings of the IEEE conference on computer vision and pattern
  recognition. pp. 839--847 (2017)

\bibitem{wang2020training}
Wang, Z., Hu, G., Hu, Q.: Training noise-robust deep neural networks via
  meta-learning. In: Proceedings of the IEEE/CVF conference on computer vision
  and pattern recognition. pp. 4524--4533 (2020)

\bibitem{wei2020combating}
Wei, H., Feng, L., Chen, X., An, B.: Combating noisy labels by agreement: A
  joint training method with co-regularization. In: Proceedings of the IEEE/CVF
  Conference on Computer Vision and Pattern Recognition. pp. 13726--13735
  (2020)

\bibitem{xiao2015learning}
Xiao, T., Xia, T., Yang, Y., Huang, C., Wang, X.: Learning from massive noisy
  labeled data for image classification. In: Proceedings of the IEEE conference
  on computer vision and pattern recognition. pp. 2691--2699 (2015)

\bibitem{xu2021faster}
Xu, Y., Zhu, L., Jiang, L., Yang, Y.: Faster meta update strategy for
  noise-robust deep learning. In: Proceedings of the IEEE/CVF Conference on
  Computer Vision and Pattern Recognition. pp. 144--153 (2021)

\bibitem{yi2019probabilistic}
Yi, K., Wu, J.: Probabilistic end-to-end noise correction for learning with
  noisy labels. In: Proceedings of the IEEE/CVF Conference on Computer Vision
  and Pattern Recognition. pp. 7017--7025 (2019)

\bibitem{zhang2018mixup}
Zhang, H., Cisse, M., Dauphin, Y.N., Lopez-Paz, D.: mixup: Beyond empirical
  risk minimization. In: International Conference on Learning Representations
  (2018)

\end{thebibliography}
	%Author, F.: Article title. Journal \textbf{2}(5), 99--110 (2016)
	%Author, F., Author, S.: Title of a proceedings paper. In: Editor,
	%F., Editor, S. (eds.) CONFERENCE 2016, LNCS, vol. 9999, pp. 1--13.
	%Springer, Heidelberg (2016). \doi{10.10007/1234567890}
	%Author, F., Author, S., Author, T.: Book title. 2nd edn. Publisher,
	%Author, A.-B.: Contribution title. In: 9th International Proceedings
	%on Proceedings, pp. 1--2. Publisher, Location (2010)
	%LNCS Homepage, \url{http://www.springer.com/lncs}. Last accessed 4
	%\end{thebibliography}
\end{document}